# Researchers eye-view of sarcasm detection in social media textual content

1st Swapnil Mane
*Department of Computer engineering
and Information technology*
*College of Engineering*
Pune, India
maness19.comp@coep.ac.in

2nd Vaibhav Khatavkar
*Department of Computer engineering
and Information technology*
*College of Engineering*
Pune, India
vkk.comp@coep.ac.in

*Abstract*—*The enormous use of sarcastic text in all forms of communication in social media will have a physiological effect on target users. Each user has a different approach to misusing and recognizing sarcasm. Sarcasm detection is difficult even for users, and this will depend on many things such as perspective, context, special symbols. So, that will be a challenging task for machines to differentiate sarcastic sentences from non-sarcastic sentences. There are no exact rules based on which model will accurately detect sarcasm from many text corpus in the current situation. So, one needs to focus on optimistic and forthcoming approaches in the sarcasm detection domain. This paper discusses various sarcasm detection techniques and concludes with some approaches, related datasets with optimal features, and the researcher's challenges.*

## I. Introduction

Sarcasm identification from text data is generally a more difficult task than sentiment identification or any other text classification task. Although the definition of 'sarcasm' from Oxford's dictionary is "A method of applying words that are the inverse of what you mean to be offensive to somebody or to make fun of them." Definition from the dictionary of Cambridge is "The use of expressions that mean the inverse of what they say, made to hurt someone's emotions or to criticize something." In general, the definition of sarcasm and metaphor is by Skalicky and Crossle [9] as 'Sarcasm' is "a form of indirect language." While someone is being sarcastic, they mean something that differs from what they said". And 'Metaphor' is "a contrast among two things to help explain something individually."

Sarcasm detection from textual data has its application in nearly every feature of our daily life, e.g., identifying user's actual attitude which is positively funny and negatively nasty, automatic feedback systems, developing and transforming business or marketing approaches according to the sarcastic response of customers, ameliorate user computer interactions, catching culprit or terrorists from investigating the sarcastic reaction of peoples after a crime or terrorist attack, achieving the contextual performance of chatbots and, etc.

Besides the growth of social media networks, the users obtain and expose their sarcastic feelings/emotions more confidently, regularly, and conveniently through their social media activities. Continuously increasing the popularity of videos, images, and audio elements, the text is still the most traditional kind of conversation in social media. The user conveys their sarcastic feelings through social media posts similar to Facebook and Twitter status, comments on Instagram and Facebook posts, Reddit comments, tweets, micro-blogs, amazon product reviews. Exploring these text corpus and identifying sarcasm from their words and semantic context is quite challenging. Sarcasm detection from text corpus has been an encouraging research topic for over a year, and essential efforts attempt to develop a precise automated system to detect accurate human sarcasm attitudes from the text corpus.

Sarcasm recognition is the procedure of detecting a human's sarcastic feelings by merely depending on his thoughts and interpretation, by using natural language processing(NLP) and machine learning(ML) techniques. Many literary works have shown invented methods for sarcasm detection, namely Tony and Aniruddha [1] have detected sarcasm with neural network semantic models composed with CNN, LSTM, and deep learning networks for Twitter. Shereen Oraby et al. [2] have applied linguistic pattern learners with weakly supervised and qualitative analysis of the linguistic dissimilarity in classes of online debate forums dialogues. Ellen Riloff et al. [3] developed a sarcasm recognizer to identify the sentiment-wise, sarcastic statement. For this model, they used a bootstrapping algorithm on Reddit Corpus (SARC).

Detection of sarcastic emotions correctly is an arduous task due to their irony and ambivalence. Different statements may express the same sarcastic opinion in multiple ways, and that statement may or may not be sarcastic. Various semantic or syntactic statements may represent identical sarcastic statements in many semantic or syntactic ways. Sarcastic emotions may differ based on the situation, history, culture, and human personal parameters. Identifying a user's sarcasm from an overall part of the text, his or her conversational audio or facial gesture or expression is complex even for a human itself. Considering the detection of sarcasm automatically by machine is a complicated task. Many researchers have been working on sarcasm identification from several sets of data inputs, e.g., text, video, image, and audio. Sarcasm detection can be performed based on various languages and multiple formats of data, such as video [4], [5] has built a model for the Czech language, an eastern German language to detect sarcasm, text [6], and text with emoji [7]. For the sarcasm detection, researchers have used time-series social media data [8] introduced approximately 300 sarcastic time-series tweets from Twitter, further detected by monolingual machine translation. Felbo et al. [7] captured tweets with various emojis for detecting the sentiment of the tweet, sarcasm of tweet, and emotions of tweets, which is to improve the weight of particular tweets and the efficiency of the model for detection. Castro et al. [4] introduced a new dataset gathered from famous TV shows and multimodel sarcasm detection, reducing sarcasm detection's error rate using multi-model information.

Existing work has explored their base-level features, approaches, and challenges for enhancing the performance. This paper attempts to incorporate the recent research works that focused on sarcasm detection in text corpus or text documents. The rest of the paper is organized as follows, the analysis of sarcasm is described in section 2, the dataset used by researchers in section 3, optimal features for sarcasm identification in section 4, in section 5, researchers' techniques and approaches and challenges over the recent research of sarcasm detection in section 6. Finally, we conclude our work with future work.

## II. Sarcasm analysis

People are expressing their feelings in multiple ways. Some of the standard methods are speech, writing, body language, facial expression, etc. Considerably, suppose to take the text to detect and analyze sarcasm from any content, which needs to design a specific approach. Figure 1 gives genres of comedy in various forms. Nearly all genres have direct sentiment or emotion, excluding sarcasm and Irony.

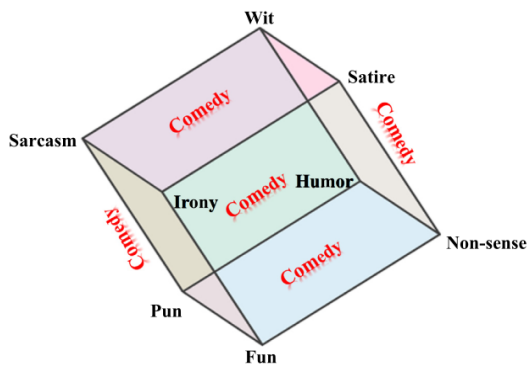

Figure 1: Genres of comedy [31]

### A. Sarcasm Evaluation

Usually, sarcasm is used for humorous purposes or to mock or annoy someone. Boyle [62] ambivalence [63] may help to achieve irony, although it is not mandatory as ironic. Partridge [64] It is noticeable in body language, spoken words, mainly depending on the context of the content. The 'Sarcasm' word introduced in English in 1579, annotated to shepherds Calendar by Edmund Spenser. Gibbs [56] proved the sarcasm was memorized much better than literal uses or typical sentences of the same expression of non-sarcastic equivalents. This analysis illustrated that ease of processing and thought for sarcastic response depends crucially upon which explicitly a user statement echoes either the addresses, opinions, or previous statement. Interest in sarcasm appears from literary theorists, linguistic and philosophical, that concern a rationalistic record of the circumstances included in understanding sarcasm. The usual common way, which we call the Standard Pragmatic Model, aims at a listener's necessity to first analyze an expression's complete literal understanding before deriving its nonliteral. sarcastic meaning [53][54][57], related to the way indirect requests, idioms, and metaphorical utterances are interpreted [58], [55] and [57]. The sarcasm and irony are mainly formed from an extensive range of situational, semantic, pragmatic, and lexical cues. In 2000 [61] focused on hyperbola and rhetorical questions with distinct types of systematic lexical cues that define sarcasm classes.

### B. Sarcasm Detection

Researchers have worked on finding patterns and recognizing contextual data. For retrieving sarcastic statements using lexical-syntactic cues along that [2] has applied weakly supervised linguistic pattern learner and qualitative analysis of the linguistic differences in classes, it is tested on online debate forums dialogue. Rolandos Potamias et al. [46] proposed a pre-trained transformer-based architecture that will be further enhanced with the recurrent convolutional neural network. This hybrid neural architecture is tested on four benchmark sarcastic datasets. It will minimize preprocessing and help to extract features.

Aboobaker and Ilavarasan [52] give the general architecture of sarcasm detection, existing systems for various sarcasm types. They also provide many of the issues, challenges as [42] developed the CRA (Contextual Response Augmentation) method that can deal with varying input and output formats and deal with variable context length effectively. That has been extended by [51] developed a system, namely eCVM, for determining context phrases and their importance, for that they use context vectors and page ranking algorithms. The context phrase has an essential role in clustering the documents that are similar in context. Focusing on contextual information, [43] has built a C-net for sequentially extracting contextual information of sentences to detect sarcasm. Also, the C-net model is showcased in the sarcasm detection shared task of CodeLab.

## C. Types of Sarcasm

In social media, users interact with each other through video, audio, and text. Social media will continuously produce such massive data in a minute. In social media, some users are sarcastically insulting some users publicly. Sarcasm is commonly used as an element of humor or as a means of insult. The following are indicative examples of sarcasm:

- Irony: The statement and in that situation (action) are contradictory to what one presumes. There is no relation between irony and sarcasm, except irony can be sarcastic.
- Flattery: A flattering compliment or speech; excessive, insincere praise.
- Insult: The insult is a statement and expression which is too disrespectful or rude. Insults may be accidental and intentional. The insult may be genuine, but at the same time, it isn't polite.
- Passive aggression: This is an indirect resistance to the demand of others.
- Humor: Sarcasm is a foundational type of humor—the tendency of sentences to provoke laughter.
- This is an overview of sarcasm in general sentences. But in social media such as Twitter, Facebook, Instagram, etc., have limitations of characters or users who want to express their feelings shortly so that users use emoji, special symbols, and interjection words. For this, [60] showcased several sarcastic statements for social media (Table I).

TABLE I. TYPE OF SARCASTIC STATEMENTS FOR SOCIAL MEDIA.

| T1 | "Sarcasm is a contradiction between positive sentiment and negative situation." |
|---|---|
| T2 | "Sarcasm is a contradiction between negative sentiment and positive situation." |
| T3 | "Statements that start with an interjection word." |
| T4 | "Sarcasm is a contradiction between likes and dislikes." |
| T5 | "Sarcasm is a contradiction between statement and the universal facts." |
| T6 | "Sarcasm is a contradiction between statement and its temporal facts." |
| T7 | "Positive statement that contains a word and its antonym pair." |

[60] Santosh Kumar et al.

## III. DATASETS USED FOR SARCASM DETECTION:

Dataset mainly used for sarcasm detection is publicly available data such as amazon product reviews, Reddit sarcastic comment corpus, sarcastic tweets based on hashtags. Mikhail Khodak et al. [6] annotated a large corpus for sarcasm detection, and this is a Reddit corpus with about 1.3 million sarcastic comments. Silviu Oprea and Walid Magdy [16], to identify the sarcasm, found out the restrictions in the method of labeling. They developed a dataset as i-sarcasm, where labeling is done with their respective author in tweets. The state of art method gives the low performance for this dataset, so they aim to encourage the researcher to work on Arthur labeled dataset, not with an assumed label. Sayed Salim et al. [17] created his Twitter sarcasm dataset and improved sarcasm detection performance using RNN, LSTM, and Word Embedding models.

The statistics of datasets used by researchers have shown in figure 2. Most of the researchers have used the Twitter dataset for sarcasm detection, Considering the five publicly available datasets, i.e., Facebook, Amazon product reviews, Reddit comments, and Twitter dataset. In contrast, very few researchers have worked on the Facebook dataset.

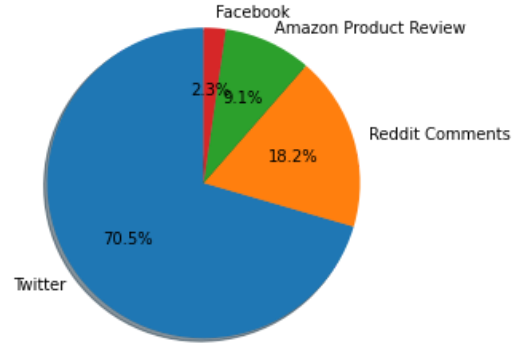

Figure 2: Statistics of datasets are used by researchers.

The information about publicly available datasets is shown in Table II. The sarcastic Facebook data is not available publicly. The data of twitter varies from individual researcher to researcher. Excepting Twitter data, all datasets are benchmark datasets for sarcasm detection.

TABLE II. DETAILS OF PUBLICLY AVAILABLE DATASETS FOR SARCASM DETECTION

| Dataset Name | Documents | Sarcastic Documents | Non-Sarcastic Documents | Average words in each document | Size |
|---|---|---|---|---|---|
| Twitter | 89536 | 22786 | 66750 | 15 | 9.75 MB |
| Reddit comments | 1010826 | 505413 | 505413 | 10 | 255.3 MB |
| Amazon Product Review | 1245 | 435 | 810 | 234 | 1.8 MB |
| News headlines | 28619 | 13634 | 14985 | 10 | 6.1 MB |

## IV. OPTIMISTIC FEATURES

The vital role of sarcasm detection is helpful to feature extraction. Ahuja and Sharma [11] have reviewed sarcasm detection and discovered that contextual data is essential. Also, they grouped various features, selection techniques, datasets. The three methodologies of sarcasm detections have been introduced based on Sentiment-Related Features, Syntactic and Semantic-Based, Punctuation Based, Pattern-Based, Short Text, Long Text, Long Text, Conversational Text, Other Miscellaneous Datasets, Rule-Based Technique, Feature Sets, and Deep Learning Methods. Roberto González-Ibáñez et al. [18] investigated features for automatically identifying sarcastic utterance in twitter corpus, that has linguistic and pragmatic factors considered in features.

Aditya Joshi et al. [19] designed contextual Incongruity features for sarcasm detection. That will improve sarcasm detection efficiency. Abhijit Mishra et al. [20] designed linguistic and stylistic features for sarcasm detection. Using these features, sarcasm detection performance has improved by nearly about 3.7% F-Score on Amazon, Twitter, and website quote corpus. Santosh Bharti et al. [10] introduced an algorithm to detect sarcasm sentences from the Telugu corpus. They achieve 94 % accuracy using their proposed algorithm with hyperbolic features such as exclamation marks, intensifier, question marks, and interjection.

They have tried to overcome the lack of resources for the languages such as Hindi, Telugu, Tamil, Arabic, etc. Bouazizi and Ohtsuki [22] propose sets of features based on patterns of tweets that cover different types of sarcasm. They detect sarcastic tweets from the Twitter corpus and achieve good results. Kolhe and Sonawane [12] contributed a technique for extracting features and selecting optimal features with the help of the T-test as an assessment method of term co-occurrence. Optimal features will additionally use this feature for defining the heuristic scale to identify the scope of sarcasm in tweets. Feature extraction for the sarcasm detection has several traditional ways, some of those elaborate below:

- N-gram: The 'N' in the N-gram approach represents a number, and it means how many words there are in one gram. In bigram is a sequence of two words. With this n-grams approach, our previous problem is a nonissue. So, it is not restricted to unigrams or tri-gram, but can be trigrams, four-grams, five-grams, whatever we think is necessary for our specific document classification task. This n-gram model is integrated into most document classification tasks, and most of the time, it boosts accuracy.

- Hashtags: A hashtag is a combination of letters, numbers, or symbols prefixed by a hash symbol, which is mandatory. Hashtags are generally used in social media such as Twitter, Facebook, and other microblogging sites. A hashtag is nothing but user-generated tagging that helps other users easily find messages with specific topics or content. So, hashtags are primarily used in the classification of social media content.

- POS-tagger: Part of Speech tagging (POS) is assigned to each word in a list using context. This is useful because the same word with a different part of speech can have two completely different meanings.

- For example, if we have two sentences ['A plane can fly' and 'There is a fly in the room'], it would be essential to define 'fly' and 'fly' correctly to determine how the two sentences are related.

- Tagging words by part of speech allows chunking. An important note is that the tagger should do POS tagging straight after tokenization and before any words are removed, that sentence structure is preserved, and it is more obvious what part of speech the word belongs to.

- Syntactic: Syntactic features denote syntax-related features that require an analysis of the document's grammatical structure to be extracted.

- Sentiment: To determine the overall opinion of the document. Sentiment analysis at the document level assumes that each document expresses views on a single entity used for additional features.

- Word Embedding: Word embedding is used in various applications of NLP. An ample of algorithms are available nowadays which are used according to their usage. Word2vec, Glove, Fast text, Bert, etc., are some of the renowned embedding techniques used by various researchers. Word embedding means mapping a word into some numerical value like a vector having multiple dimensions. The vector assigned to any of the words depends upon the context of the word. So, a vector can have several dimensions such as 50,100, 200, 300, etc. In most of the case, embedding is the first step in a neural network.

V. APPROACHES AND TECHNIQUES

As we focus on the context of social media data, researchers have done some work on the sentiment of context as a feature for sarcasm detection. Tony and Aniruddha [13] focused on the detection of sarcasm in tweets. Tony and Aniruddha prove that the new post of Twitter has a cue for irony based on user tweets expressing their feelings over a top position. That is the contextual topic of the post itself. Ameeta Agrawal et al. [14] investigated whether emotions mattered in sarcasm detection or not by using two datasets of different genres and sizes. This proposed approach significantly benefits datasets with limited labeled data and longer instances of text which are generally more challenging to annotate. Navonil Majumder et al. [15] showed that sentiment and sarcasm tasks are correlated and developed a multitask model that is learning-based implemented using a deep neural network to enhance sentiment accuracy sarcasm tasks on the dataset 16. Ellen Riloff et al. [3] developed a sarcasm recognizer to identify the sentiment-wise, sarcastic statement. For this model, they used a bootstrapping algorithm on Reddit Corpus (SARC).

Bjarke Felbo et al. [7] captured tweets with various emojis for detecting the sentiment of the tweet, sarcasm of tweet, and emotions of tweets, which is to improve the weight of particular tweets and the efficiency of the model for detection. Taha Ataei et al. [23] proposed a response sarcasm detection model. They captured response and contextual dialogue sequences using a sentiment analysis approach that is an aspect-based technique and BERT. Ibrahim Abu-Farha and Magdy [24] prove that the sarcastic content has affected sentiment detection, reducing its performance. Ibrahim uses BiLSTM for sarcasm detection in Arabic tweets. Savini and Caragea [25] explored the multitask learning framework. It uses sentiment classification as an auxiliary task to inform primary studies of sarcasm detection. In that, they exploit the relation between sarcasm and negative sentiment that the message conveys. Meishan Zhang et al. [26] investigated whether the tweets are sarcastic or not by using the tweet's semantic and syntactic information. For detection, they use bi-directional gated recurrent neural networks (RNN), and contextual features are automatically captured by pooling neural networks from history tweets. Lu Ren et al. [27] proposed multilevel memory network sentiment semantics (MMNSS). For this, they extract the features for detecting sarcasm. In

Ren's model, for extracting sentiment semantics, they use memory networks of its first level and extract the difference between the situation and sentiment semantics in each sentence or expression. They use the second level of the memory network. For improving the memory network, they used CNN deep neural network models with an absence of local information.

Traditional detection techniques for sarcasm are not efficient for getting an adequate performance with an original social media corpus as a feature. So, researchers focused more on design feature vectors to detect sarcasm and develop a model to detect it. Kalai Vani et al. [28] categorized and identified the context of conversation and audience that a response or repliers user is used to detecting sarcasm in Twitter and Reddit dataset using RNN-LSTM and BERT. Suzana Ilić et al. [29] Proposed model is a character label representation of vectors for each word using ELMo. This proposed model is tested on seven datasets formed from three data sources, six of which give a state-of-the-art performance. Sahil Jain et al. [30] detects sarcasm reviews from amazon product reviews using traditional state-of-art methods from Sahil observed that SVM has the highest accuracy than other methods. Jain Deepak et al. [31] proposed a model to detect sarcasm from the Hinglish (Hindi + English) corpus of Twitter. They use a hybrid of BiLSTM and that attention layer of softmax and CNN for sarcasm detection in time series tweets. Nikhil Jaiswal [32] investigated several types of models to represent pre-trained languages such as BERT, RoBARTa, etc. Using ensemble model performance of classification is improved on Twitter sarcasm dataset. Akshay et al. [33] for sarcasm detection uses machine learning techniques as BERT and GloVe embedding for Twitter. Xiang Dong et al. [34] gives a sarcasm detection model based on transformation. It establishes a strong baseline for a target-oriented model compared to the context-aware models that show significant improvements for Reddit and Twitter data. Dark under Mayur [35] proposed hyperparameter optimization LSTM using genetic algorithm, on that, they applied CNN. Arup Baruah et al. [36] present sarcasm detection performance using SVM, BiLSTM, and BERT classifiers on Twitter and Reddit sarcasm datasets. They prove BERT classifiers are highly efficient than BiLSTM and SVM classifiers for sarcasm detection. Kartikey and Tanvi [37] used the RoBARTalarge method and word embedding model with three distinguishing inputs: Context-Response (Separated) Context-Response, and Response-only for improving the performance of sarcasm detection for Twitter and Reddit datasets.

Multi-head attention (MH A) based bidirectional long-short memory (BiLSTM) model for detecting sarcastic sentences, this is proposed by [38]. This proposed system will enhance the performance of BiLST. Diao et al. [39] introduced a Question Answering (QA) system using a multidimensional model to detect the sarcastic response, and it overcomes ambiguity of sarcasm by multidimensional representation. They built contextual information for conversation. A deep memory QA network depends on BiLSTM with an attention network to discover sarcasm sentences. Yi Tay et al. [40] proposes the attention-based neural network model in between instead of across. Tested on six benchmark datasets are Reddit, Twitter, and Internet Argument Corpus.

The summarization of each research approach with their performances for sarcasm detection is described in the following tables. The tables are formed based on techniques used in the approach. Table III describes machine learning algorithms used by researchers for sarcasm detection. Table IV describes deep neural network algorithms used by researchers for sarcasm detection. Table V describes deep and recurrent neural network algorithms used by researchers for sarcasm detection. We observed that most of the researchers had used the most Long short-term memory as an artificial neural network for sarcasm detection.

Researchers have worked on text, audio speech, video data. Much research has been done on Twitter time-series data, amazon product reviews, and Reddit comments. This dataset is used for sarcasm detection. For features, they used word embedding vectors, syntactic pattern, sentiment, POS-tagger, hashtags in tweets, and N-gram; these are the most trending feature selection techniques in sarcasm detection. Researchers have used various neural network models for sarcasm detection, namely RNN, LSTM, CNN, etc. But they have not achieved good results on benchmark datasets because of the lack of contextual information for sarcastic or non-sarcastic classes. The table is described at the end of the paper.

## VI. CHALLENGES

The system built using recent approaches provides better results than previous works, including traditional methods. By using neural networks, the performance of sarcasm detection has to be improved by considering several features. Most of the researchers work on a commonly available social media dataset, namely Twitter data. Despite that, no such models will significantly work, including the contextual information of users' data and modeling. This is a remarkable research gap in the domain of sarcasm detection. In the detection of sarcasm, there is a lack of research for general models that will work, including the context of data and for several sources of datasets that have multiple forms.

Considering recent research, we summarized some challenges in sarcasm detection, which will enhance sarcasm detection performance.

- A more extensive corpus of different domains: The dataset for sarcasm detection is not standardized in the current situation. That is, the data is not too large and acceptable for enhancing the performance of models. As an observation, the extent of work in the sarcasm detection domain is limited.

- User's previous information is a feature: Detecting sarcasm is a critical task for machines and even humans. So, we can capture the user's past information or user's profile, and this is a piece of essential information required for the detection of sarcasm.

- Context of data: The role of context is more essential for improving existing systems' performance because context is more effective in detecting sarcasm rather than a pattern.

- Generalized system or approach: According to recent research, a generalized system or approach is saturation in the system's performance as the text corpus is more grown-up than a specific optimized amount. In that case, the system will go into overfitting or underfit problems, for that necessary to obtain the balance between them by combining datasets, changing the model complexity.

- The sarcastic document's actual response: After detecting the sarcastic document, need to find out the document's sentiment. Translating the sarcastic document to the user's actual response is a critical challenge for the upcoming researcher.

## VII. Conclusion and future work

This paper discovers a birds-eye view of all aspects of sarcasm detection. It also discusses which datasets, features, and respective techniques in recent researchers have been mostly used for sarcasm detection. It is found that multi-instance logistic regression [20], random forest [22] and voting classifier [41] machine learning algorithms have efficient performance than other traditional machine learning algorithms. In the deep learning algorithm, [46], the RCNN-Roberta model improves accuracy more efficiently. Sonawane and Kolhe [12] proposed that LSTM is much more accurate for sarcasm detection compared to the other approaches. Many researchers have contributed a lot to improve the performance of sarcasm detection. But at some point, their proposed method is not enhancing performance due to a lack of sarcasm knowledge. This paper discusses sarcasm detection challenges considering the domain, contextual information, general approaches, and translation to actual response of sarcasm. Researchers may work on the machine translation to translate the sarcastic sentence to the actual response in future work. Sarcasm detection aims to achieve the document's sentiment, so machine translation is necessary to accomplish the document's actual opinion.

The details of Table names:
D1: Twitter dataset
D2: Reddit dataset
D3: Facebook dataset
D4: Other's dataset
F1: Embedding features
F2: Hashtags features
F3: N-gram features
F4: POS-tagger features
F5: Syntactic features
F6: Sentiment features
F7: Other features
Note: F- F1-score, Acc- Accuracy, P- Precision

TABLE III. SUMMARIZE KEY POINTS AND PERFORMANCE OF RECENT RESEARCH DONE IN THIS DOMAIN USING MACHINE LEARNING ALGORITHMS.

| Reference | Year | Dataset | | | | Features | | | | | | | Models | Accuracy |
| --- | --- | --- | --- | --- | --- | --- | --- | --- | --- | --- | --- | --- | --- | --- |
| | | D1 | D2 | D3 | D4 | F1 | F2 | F3 | F4 | F5 | F6 | F7 | | |
| Bouazizi And Otsuki [22] | 2016 | ✓ | | | | | | ✓ | | ✓ | ✓ | | RF, SVM, KNN Max. Ent. | F: 81.3(RF) 33.8(SVM), 79.6(KNN), 74.8(Max. Ent.) |
| Rahul Gupta et al. [41] | 2020 | ✓ | | ✓ | ✓ | | | ✓ | | | ✓ | | SVM, Voting classifier | Acc: 74.59(SVM), 83.53(voting classifier) |
| Akshay et al. [33] | 2020 | ✓ | | | | ✓ | | | | | | | Linear SVM, Logistic Regression, Gaussian Naive Bayes, Random Forest | F: 67.9(LinSVM),69(LogReg) |
| Santosh Bharti et al. [44] | 2019 | | | ✓ | | | | | ✓ | ✓ | | | Naive Bayes, SVM, Decision Tree, KNN, Random Forest, Adaboost | Acc: 85 |
| Pawar and Bhingarkar [45] | 2020 | ✓ | | | | | | | ✓ | ✓ | | | Random forest classifier, SVM, KNN | Acc: 81 (RF), 74(SVM) 58(KNN) F: 79 (RF) 22.15(SVM) 66.8(KNN) |
| Shubhodip Saha et. al. [47] | 2017 | ✓ | | | | | | ✓ | ✓ | ✓ | | | Naïve Bayes, SVM | Acc: 65.2(NB), 60.1(SVM) |
| Tomáš Ptáček et. al. [5] | 2014 | ✓ | | | | | | ✓ | ✓ | ✓ | | | Maximum Entropy (MaxEnt) and Support Vector Machine | F: 94.7(English), 58.2 (Czech) |
| Abhijit Mishra et al. [20] | 2017 | ✓ | | | ✓ | | ✓ | | | ✓ | ✓ | ✓ | Multi Instance Logistic Regression | P: 84 |
| Roberto González-Ibáñez et. al.[18] | 2011 | ✓ | | | | | ✓ | | | ✓ | | | sequential minimal optimization and logistic regression. | Acc: 78(SMO), 75(LogR) |

TABLE IV. SUMMARIZE KEY POINTS AND PERFORMANCE OF RECENT RESEARCH DONE IN THIS DOMAIN USING DEEP NEURAL NETWORK ALGORITHMS.

| Reference | Year | Dataset | | | | Features | | | | | | | Models | Accuracy D1 |
| --- | --- | --- | --- | --- | --- | --- | --- | --- | --- | --- | --- | --- | --- | --- |
| | | D1 | D2 | D3 | D4 | F1 | F2 | D1 | D2 | D3 | D4 | F7 | | |
| Amit Kumar et al. [43] | 2020 | ✓ | ✓ | | | ✓ | | | | | ✓ | | C-net (includes 3 models) | F: 71(twitter), 66.3 (reddit) |
| Nikhil Jaiswal [32] | 2020 | ✓ | | | | ✓ | ✓ | | | | | | ensemble model, Embeddings from Language Models, Universal Sentence Encoder, Bidirectional Encoder Representations from Transformers (BERT), Robustly Optimized BERT Approach (RoBERTa) | F: 79 |
| Rolandos Potamias et. al. [46] | 2020 | | ✓ | | ✓ | ✓ | | | | | ✓ | | RCNN-RoBERTa model | Acc:90, Cosin:81, mse: 1.45 |
| Aditya Joshi et al. [19] | 2015 | ✓ | | | ✓ | | | | ✓ | ✓ | ✓ | | LibSVM with RBF kernel (Chang and Lin, 2011), and report average 5-fold cross-validation values | F: 88.76 |
| Silvio Amir et al. [48] | 2016 | ✓ | | | | | ✓ | | | | ✓ | | SHALLOW CUE-CNN | Acc: 87.2 |

TABLE V. SUMMARIZE KEY POINTS AND PERFORMANCE OF RECENT RESEARCH DONE IN THIS DOMAIN USING DEEP AND RECURRENT NEURAL NETWORK ALGORITHMS.

| Reference | Year | Dataset | | | | Features | | | | | | Models | Accuracy |
| --- | --- | --- | --- | --- | --- | --- | --- | --- | --- | --- | --- | --- | --- |
| | | D1 | D2 | D3 | D4 | F1 | F2 | D1 | D2 | D3 | D4 | | |
| darkunde mayur ashok et. al [35] | 2020 | ✓ | | | | | ✓ | | | | | LSTM, CNN | Acc: 85(CNN), 90(LSTM-CN),93(proposed model) |
| sayed salim et. al. [17] | 2020 | ✓ | | | | ✓ | ✓ | | | | | RNN, LSTM | Acc: 88 |
| Xiang Dong et al. [34] | 2020 | ✓ | ✓ | | | ✓ | ✓ | | | | | BERT-Large, RoBERTa-Large and ALBERT- xxLarge | Acc: 63.1 (twitter), 74 (reddit) |

| Author | Year | C1 | C2 | C3 | C4 | C5 | C6 | C7 | C8 | Model | Result |
|---|---|---|---|---|---|---|---|---|---|---|---|
| kartikey and tanvi [37] | 2020 | ✓ | ✓ | | ✓ | | | | ✓ | RoBERTalarge | F: improvement of 5.13 |
| hanky ohl et. al.[42] | 2020 | ✓ | ✓ | | ✓ | | | | ✓ | BERT(large-cased): 24-layer, 1024-hidden and 16-heads. BiLSTM: 2-layer, 1024-hidden and 0.25-dropout. NeXt VLAD: 8-groups, 4-expansion, 128- number of clusters and 512-cluster size. | F: Around 73.13 |
| taha ataei et. al. [23] | 2020 | ✓ | ✓ | | ✓ | ✓ | ✓ | | | method - NB-SVM, Transformers - BERT-base-cased, Bi-GRU-CNN+BiLSTM-CNN, Aspect-based - LCF-BERT, IAN, BERT-AEN | F: 73 (Twitter), 73.4 (Reddit) |
| kalaivani et. al. [28] | 2020 | ✓ | ✓ | | ✓ | | | | | RNN-LSTM and BERT | F: 72.2 (Twitter), 67.9 (Reddit) |
| Arup Baruah et al. [36] | 2020 | ✓ | ✓ | | ✓ | | | ✓ | | BERT, BiLSTM and SVM classifiers | F: 74.3 (Twitter), 65.8 (Reddit) |
| Ibrahim and Walid [24] | 2020 | ✓ | | | | | | | ✓ | BiLSTM | F: 46 |
| Sonawane and Kolhe [12] | 2019 | ✓ | | | | | | | ✓ | TCSD | Acc: 95 |
| Savini and Caragea [25] | 2020 | | | ✓ | ✓ | | | | ✓ | BiLSTM + MTL | Acc: 76 |
| Ameeta Agrawal et al. [14] | 2020 | | | ✓ | ✓ | | | ✓ | ✓ | LSTM with emotion vector but no chunking (Emotion No Chunk): | Acc: 65 |
| Avinash Kumar et al. [38] | 2020 | | | ✓ | ✓ | | | ✓ | ✓ | MHA - BiLSTM | F: 77.48 |
| YUFENG DIAO et. al. [39] | 2020 | | | ✓ | ✓ | | ✓ | | | MQA | F: 76.2 |
| Deepak Jain et. al. [31] | 2020 | ✓ | | | ✓ | | ✓ | ✓ | | BiLSTM -> attentionn layer -> CNN | Acc: 92.71 and F: 89.05 |
| Rolandos Potamias et. al. [46] | 2020 | | ✓ | ✓ | ✓ | | | | ✓ | RCNN-RoBERTa model | Acc:90, Cosin:81, mse: 1.45 |
| Lu Ren et. al. [27] | 2020 | ✓ | | ✓ | ✓ | ✓ | | | ✓ | MMNSS (Multi-level Memory Network sentiment semantic) | F: 87.13 |
| Ghosh and Veale [1] | 2016 | ✓ | | | | | ✓ | ✓ | ✓ | CNN + LSTM + DNN (without dropout) | F: 92 |
| Ghosh and Veale [13] | 2017 | ✓ | | | ✓ | ✓ | | | | CNN1 + BLSTM1 and CNN2 + BLSTM2 | F: 90 |
| Ellen Riloff et. al. [3] | 2013 | ✓ | | | | ✓ | ✓ | | ✓ | Bootstrapped | F: .51 |
| Navonil Majumder et. al. [15] | 2019 | | | ✓ | | | | | ✓ | multi task GRU | F: 90.67 |
| Bjarke Felbo et. al. [7] | 2017 | ✓ | | | ✓ | | | | ✓ | DeepMoji model with smaller LSTM layers and a bag-of-words classifier | Acc: 82.4 |